\pdfoutput=1

\documentclass[twocolumn,11pt]{article}

\usepackage[preprint]{acl}

\usepackage{times}
\usepackage{latexsym}

\usepackage[T1]{fontenc}

\usepackage[utf8]{inputenc}

\usepackage{microtype}

\usepackage{inconsolata}

\usepackage{graphicx}

\usepackage{booktabs}
\usepackage{multirow}
\usepackage{float}
\usepackage{caption}
\usepackage[normalem]{ulem}
\useunder{\uline}{\ul}{}

\usepackage{hyphenat}
\usepackage{fontawesome5}
\usepackage[all]{nowidow}

\title{Domain-Specific Translation with\\
Open-Source Large Language Models:\\
Resource-Oriented Analysis}

\author{Aman Kassahun Wassie*
  \and
  Mahdi Molaei*
  \and
  Yasmin Moslem*\textsuperscript{\tiny\faStar[regular]}
  }

\date{}

\begin{document}
\maketitle

\def\thefootnote{*}\footnotetext{The three authors contributed equally to this work.}\def\thefootnote{\arabic{footnote}}
\def\thefootnote{{\scalebox{0.5}{\faStar[regular]}}}
\footnotetext{Correspondence: \url{yasmin [at] machinetranslation.io}}\def\thefootnote{\arabic{footnote}}

\begin{abstract}
\nohyphens{
In this work, we compare the domain-specific translation performance of open-source autoregressive decoder-only large language models (LLMs) with task-oriented machine translation (MT) models. Our experiments focus on the medical domain and cover four language directions with varied resource availability: English-to-French, English-to-Portuguese, English-to-Swahili, and Swahili-to-English. Despite recent advancements, LLMs demonstrate a significant quality gap in specialized translation compared to multilingual encoder-decoder MT models such as NLLB-200. Our results indicate that NLLB-200 3.3B outperforms all evaluated LLMs in the 7-8B parameter range across three out of the four language directions. While fine-tuning improves the performance of LLMs such as Mistral and Llama, these models still underperform compared to fine-tuned NLLB-200 3.3B models. Our findings highlight the ongoing need for specialized MT models to achieve high-quality domain-specific translation, especially in medium-resource and low-resource settings. Moreover, the superior performance of larger LLMs over their 8B variants suggests potential value in pre-training domain-specific medium-sized language models, employing targeted data selection and knowledge distillation approaches to enhance both quality and efficiency in specialized translation tasks.
 }

\end{abstract}

\section{Introduction}

In recent years, there have been several advancements related to employing large language models (LLMs) in diverse natural language processing (NLP) tasks, including machine translation (MT). Autoregressive decoder-only models such as BLOOM \citep{BLOOM2022}, ChatGPT \citep{Brown2020-GPT-3}, Mistral \citep{Jiang2023-Mistral}, and Llama \citep{Touvron2023-Llama2} --to mention but a few-- have been welcomed by NLP researchers and practitioners as a new paradigm, due to their outstanding in-context learning feature. As the name suggests, in-context learning enables an LLM to leverage additional context during inference to enhance its output \citep{Brown2020-GPT-3}. When it comes to MT, researchers investigated incorporating all sorts of context, such as translation pairs similar to the new source text \citep{Agrawal2023-SelectionMT,Moslem2023-AdaptiveMT,Vilar2023-PaLM-MT}, terminology \citep{Moslem2023-Terminology}, and dictionary words \citep{Ghazvininejad2023-DictionaryMT}. Usually prompting the model, i.e. providing instructions in a natural language, is sufficient without additional model fine-tuning. Nevertheless, several researchers investigated fine-tuning LLMs for MT tasks, for both sentence-level translation \citep{Alves2023-LLM-MT-Finetuning,Xu2023-Llama-Finetuning}, and adaptive translation \citep{Moslem2023-AdaptiveMT-LLM-Finetuning,Zhang2023-MT-Efficient-Finetuning,Stap2024-LLM-Finetuning}.

Due to data privacy and security concerns, several businesses and organizations prefer to use open-source LLMs in their NLP tasks. However, despite recent advances, there are still several challenges for using LLMs for MT, including: 

\begin{itemize}
    \item clear gaps when it comes to language support, whether for the generic domain \citep{Robinson2023-ChatGPT-MT}, or for specialized domains \citep{Moslem2023-AdaptiveMT}.
    \item challenges of deploying open-source LLMs in production given their massive sizes. The size of Llama 8B is more than double the size of NLLB-200 3.3B, while Llama 70B and Llama 405B are more than 20 times and 122 times larger, respectively. 
\end{itemize}

This reopens the door to the question whether we still need conventional encoder-decoder MT models. In this work, we focus on domain-specific translation, and use NLLB-200 3.3B, both the baseline and fine-tuned models, as our main task-oriented benchmarks. NLLB-200 3.3B has demonstrated high quality comparable with strong commercial systems in diverse languages \citep{Moslem2023-AdaptiveMT}. The majority of decoder-only LLMs we explore are in the size range of 8B parameters, namely Mistral 7B \citep{Jiang2023-Mistral}, Llama-3 8B,  Llama-3.1 8B \citep{Dubey2024-Llama3}, Gemma 7B \citep{Mesnard2024-Gemma}, Qwen's QwQ-32B-Preview, and Qwen-2.5 72B \citep{Qwen2024}. Moreover, we evaluate Llama-3 70B, and Mixtral 8x7B, which is a mixture of 7B experts model, as well as the massive Llama-3.1 405B and DeepSeek \citep{DeepSeek2024}.

Even when comparing two languages with relatively higher data resources, such as French and Portuguese, we observe a considerable difference in the translation performance of these models. Moreover, for specialized translation, such as in the medical domain, many languages suffer from resource scarcity. This is clearly demonstrated by the quality of the English-to-Swahili and Swahili-to-English medical translation. Such resource differences between French, Portuguese, and Swahili are clearly reflected in their support levels by LLMs.

The main takeaways from our study include:
\begin{itemize}
    \item Domain-specific translation, e.g. in the medical domain, is challenging for general-purpose open-source LLMs, even for higher-resource languages.
    \item In only one out of the four language directions included in our study, i.e. English-to-French translation, decoder-only LLMs in the size range of 8B parameters outperform the encoder-decoder task-oriented NLLB-200 3.3B in zero-shot translation. For the other three language directions, English-to-Portuguese, English-to-Swahili, and Swahili-to-English, NLLB-200 3.3B \mbox{excels} at sentence-level translation.
    \item The massive Llama-3.1 405B outperforms most of the baseline models in the study. However, deploying Llama-3.1 405B in production can be challenging and inefficient.\footnote{Quantized versions of Llama-3.1 405B and DeepSeek V3 685B require 3-4 H100 GPUs and 8 H100 GPUs respectively for one active user. Scaling this number to simultaneously serve several users can be too expensive for most medium-sized businesses. Moreover, latency can be counterproductive in translation settings.} Hence, it can be used for offline data augmentation and knowledge distillation to smaller models.
    \item Fine-tuning task-oriented MT models such as NLLB-200 3.3B can be effective when a medium-sized, domain-specific dataset is available. In such cases, using decoder-only LLMs with around 8B parameters may lead to diminishing returns.
\end{itemize}

The next sections elaborate on our methodology and evaluation of various open-source LLMs for domain-specific, sentence-level zero-shot translation and retrieval-augmented one-shot translation, as well as fine-tuning LLMs for both use cases.

\section{Data}
\label{sec:data}

For our experiments, we used diverse publicly available datasets. We prepared our data for retrieval-augmented translation with fuzzy matches. This section elaborates on the data statistics and the process of data preparation.

\subsection{French and Portuguese}

We gathered medical data for English-to-Portuguese (EN-PT) and English-to-French (EN-FR) translations from the OPUS repository, specifically the ELRC, EMEA \citep{Tiedemann2012-OPUS}, SciELO \citep{Soares2018-SciELO}, and TICO-19 \citep{Anastasopoulos2020-TICO-19} datasets.
Following a combination of rule-based and semantic filtering techniques, we curated 2,905,418 segments for EN-PT and 303,948 segments for EN-FR.\footnote{Data preparation scripts are available at: \url{https://github.com/ymoslem/MT-Preparation/}
}

For the training process, we combined data from the aforementioned medical datasets and performed random sampling from the resulting large dataset. For each language pair, we selected 10,000 segments for a small training set, 100,000 segments for a medium-sized training set, and 10,000 segments for the test set in the medical domain. To retrieve fuzzy matches, we extracted semantically similar translation pairs from a context dataset. We sampled unique 50,000 segments for the context dataset of the small training dataset, and more 50,000 segments, making a context dataset with 100,000 for the medium training dataset (cf.~Table \ref{tab:data-sampled}).

\subsection{Swahili}

We collected medical and generic datasets for Swahili-to-English (SW-EN) and English-to-Swahili (EN-SW) models from OPUS \citep{Tiedemann2012-OPUS}. We used Global Voices and Wikimedia for the generic domain, and TICO-19 \citep{Anastasopoulos2020-TICO-19} and Wikipedia articles for the medical domain. After applying rule-based and semantic filtering, we obtained 39,415 segments for the generic domain and 3,514 segments for the medical domain.

In the generic domain, we used 10,000 segments for the test and 10,000 segments for the test context. Since the data we have for the medical domain is limited, we used only 1,000 segments for the test and 1,000 segments for the test context. For training, we mixed datasets from both the medical and generic domains. In total, we prepared 10,000 segments, with 9,000 from the generic domain and 1,000 from the medical domain. For the training context, we used all the remaining data, which includes 514 segments from the medical domain and 10,415 segments from the generic domain  (cf.~Table \ref{tab:data-sampled}).

\section{Experiments}
\label{sec:experiments}

In our experiments, we start by evaluating the baselines of various open-source models, including both autoregressive decoder-only LLMs, and the encoder-decoder NLLB-200 3.3B model, for both zero-shot and one-shot translation. Then, we fine-tune the models that show the best performance. In most cases, the LLMs chosen for fine-tuning are Mistral 7B and Llama-3 8B for their translation quality and efficiency. Following \citet{Moslem2023-AdaptiveMT},
\footnote{Code for the experiments: \url{https://github.com/ymoslem/Adaptive-MT-LLM-Fine-tuning}}
we fine-tune these LLMs using a combination of zero-shot and one-shot prompts in the medical domain. For the one-shot prompt experiments, we augment each translation pair with an additional translation example that is a fuzzy match to the source sentence. We extract fuzzy matches from a unique dataset which we refer to as the \mbox{“context dataset”} (cf.~Section \ref{sec:data}). Fuzzy matches are identified through calculating the semantic similarity between the embeddings of the source segments in the training datasets and those in the “context dataset”. The top match is selected for a one-shot prompt. Both the baseline and the fine-tuned models are then evaluated separately on zero-shot and one-shot prompts.

Fuzzy match retrieval involves three steps: (i) creating the embeddings, (ii) indexing, and (iii) retrieval. For creating the embeddings, we use Sentence-Transformers \citep{Reimers2019-SentenceTransformers} with Microsoft \textit{Multilingual-MiniLM-L12-H384}. Additionally, we tested other embedding models, and reported the results in Appendix \ref{sec:appendix-training}. For indexing, we use Faiss \citep{Johnson2019-Faiss}, a library for efficient similarity search and clustering of dense vectors, training an IndexIVFFlat index with IndexFlatL2 as a quantizer. Retrieval was done by similarity search with cosine similarity.

In our experiments, we fine-tune a range of autoregressive decoder-only models, namely: Mistral 7B, Mixtral 7x8B, Llama-3 8B, and Gemma 7B. For efficiency, we employ QLoRA, a parameter-efficient fine-tuning technique. We fine-tune the models for one epoch, as in \citet{Moslem2023-AdaptiveMT}, following the same configuration settings. As Llama-3 8B revealed the best results (cf.~Section \ref{sec:eval}), we experimented with fine-tuning it with two sizes of data sets, (i) small [s+1], which is used for fine-tuning all the other models, and (ii) and medium [m+1]. The details of these datasets can be referred to in Section \ref{sec:data}.

Moreover, we evaluated the NLLB-200 3.3B model, which is a state-of-the-art sequence-to-sequence machine translation model, based on the encoder-decoder Transformer architecture. For NLLB-200 3.3B fine-tuning, we conducted 4 experiments: (i) sentence-level fine-tuning with the small dataset [s]; (ii) sentence-level fine-tuning with the medium dataset [m]; (iii) concatenation of a fuzzy match with the new sentence pair in the small dataset [s+1]; and (iv) concatenation of a fuzzy match with the new sentence pair in the medium dataset [m+1]. The concatenated fuzzy segment and new segment are separated by both the language code and a special separator \citep{Bulte2019-fuzzy,Moslem2023-AdaptiveMT-LLM-Finetuning}. At the decoding time, for the two models that use fuzzy-match concatenation, we apply teacher forcing of the fuzzy target as a translation prefix so that the model completes the translation of the new source accordingly.\footnote{Refer to the CTranslate2's “target\_prefix” option.}

For efficient inference, we use vLLM \citep{Kwon2023-vLLM} for translation with all autoregressive models, while we use CTranslate2 for NLLB-200 3.3B, which is not supported by vLLM.\footnote{CTranslate2 supports some, but not all the autoregressive models in our experiments; hence, we decided to use vLLM for these models. Appendix \ref{sec:appendix-inference} compares scores obtained from CTranslate2 and vLLM.} Due to the massive size of Llama 3.1 405B and DeepSeek V3 685B, we had to use Activation-aware Weight Quantization (AWQ) \citep{Lin2023-AWQ}, while other baseline models do not use quantization at inference time.

We conducted our experiments with both higher-resource (French and Portuguese) and lower-resource (Swahili) languages. French and Portuguese have extensive data available, including medical datasets. However, for Swahili, due to limited data, we created a small dataset in the medical domain and a generic dataset that includes data outside the medical domain. These datasets were partitioned into training and test sets, with the remaining data designated as the context dataset (cf.~Section \ref{sec:data}).

Moreover, we ran several additional experiments to find the best set up, including embedding models, learning rates, and inference engines. Appendix \ref{sec:appendix-training}, \ref{sec:appendix-inference}, and \ref{sec:appendix-models} briefly summarize these experiments.

\begin{table*}[!ht]
\centering
\begin{small}

\begin{tabular}{@{}llllll@{}}
\toprule
\textbf{Model}                    & \textbf{Model Type}       & \textbf{Context} & \textbf{BLEU} & \textbf{chrF++} & \textbf{COMET} \\ \midrule
\multirow{8}{*}{NLLB-200 3.3B}    & \multirow{2}{*}{baseline} & zero-shot        & 41.37         & 64.10           & 59.32          \\
                              &                                      & one-shot  & 43.04          & 65.55          & 55.84          \\ \cmidrule(l){2-6} 
                              & finetuned {[}s{]}                    & zero-shot & 44.22          & 66.94          & 75.40          \\ \cmidrule(l){2-6} 
                              & finetuned {[}m{]}                    & zero-shot & 49.56          & 70.96          & 81.36          \\ \cmidrule(l){2-6} 
                              & \multirow{2}{*}{finetuned {[}s+1{]}} & zero-shot & 45.20          & 68.25          & 77.07          \\
                              &                                      & one-shot  & 47.93          & 69.15          & 77.95          \\ \cmidrule(l){2-6} 
                              & \multirow{2}{*}{finetuned {[}m+1{]}} & zero-shot & 53.13          & \textbf{73.49} & \textbf{83.83} \\
                              &                                      & one-shot  & 53.67          & 72.74          & 81.19          \\ \midrule
\multirow{4}{*}{Gemma 7B}     & \multirow{2}{*}{baseline}            & zero-shot & 26.21          & 48.31          & 33.91          \\
                              &                                      & one-shot  & 47.97          & 67.78          & 71.64          \\ \cmidrule(l){2-6} 
                              & \multirow{2}{*}{finetuned {[}s+1{]}} & zero-shot & 41.08          & 63.74          & 68.79          \\
                              &                                      & one-shot  & 41.42          & 65.87          & 48.94          \\ \midrule
\multirow{6}{*}{Llama-3 8B}   & \multirow{2}{*}{baseline}            & zero-shot & 34.40          & 58.08          & 59.36          \\
                              &                                      & one-shot  & 48.45          & 67.79          & 72.80          \\ \cmidrule(l){2-6} 
                              & \multirow{2}{*}{finetuned {[}s+1{]}} & zero-shot & 43.71          & 64.93          & 73.05          \\
                              &                                      & one-shot  & 51.11          & 69.53          & 77.88          \\ \cmidrule(l){2-6} 
                              & \multirow{2}{*}{finetuned {[}m+1{]}} & zero-shot & 49.67          & 69.15          & 77.87          \\
                              &                                      & one-shot  & \textbf{54.77} & 72.00          & 79.75          \\ \midrule
\multirow{2}{*}{Llama-3.1 8B} & \multirow{2}{*}{baseline}            & zero-shot & 33.89          & 57.67          & 58.58          \\
                              &                                      & one-shot  & 48.58          & 67.96          & 73.13          \\ \midrule
\multirow{2}{*}{Llama-3 70B}  & \multirow{2}{*}{baseline}            & zero-shot & 39.74          & 62.51          & 66.75          \\
                              &                                      & one-shot  & 51.93          & 70.57          & 76.76          \\ \midrule
\multirow{2}{*}{Llama-3.1 405B}   & \multirow{2}{*}{baseline} & zero-shot        & 42.61         & 64.92           & 69.56          \\
                              &                                      & one-shot  & 53.61          & 71.83          & 79.32          \\ \midrule
\multirow{4}{*}{Mistral 7B}   & \multirow{2}{*}{baseline}            & zero-shot & 30.61          & 54.79          & 48.56          \\
                              &                                      & one-shot  & 46.18          & 66.12          & 69.13          \\ \cmidrule(l){2-6} 
                              & \multirow{2}{*}{finetuned {[}s+1{]}} & zero-shot & 39.46          & 64.63          & 71.84          \\
                              &                                      & one-shot  & 48.40          & 69.20          & 76.64          \\ \midrule
\multirow{2}{*}{BioMistral}   & \multirow{2}{*}{baseline}            & zero-shot & 24.13          & 51.39          & 42.10          \\
                              &                                      & one-shot  & 43.41          & 63.71          & 63.81          \\ \midrule
\multirow{4}{*}{Mixtral 8x7B} & \multirow{2}{*}{baseline}            & zero-shot & 38.06          & 61.17          & 62.29          \\
                              &                                      & one-shot  & 50.98          & 69.79          & 75.57          \\ \cmidrule(l){2-6} 
                              & \multirow{2}{*}{finetuned {[}s+1{]}} & zero-shot & 44.32          & 66.19          & 75.50          \\
                              &                                      & one-shot  & 47.29          & 69.10          & 78.74          \\ \midrule
\multirow{2}{*}{Qwen QwQ 32B} & \multirow{2}{*}{baseline}            & zero-shot & 39.24          & 62.33          & 67.26          \\
                              &                                      & one-shot  & 50.39          & 69.43          & 75.52          \\ \midrule
\multirow{2}{*}{Qwen-2.5 72B} & \multirow{2}{*}{baseline}            & zero-shot & 41.27          & 63.76          & 69.82          \\
                              &                                      & one-shot  & 51.86          & 70.53          & 76.97          \\ \midrule
\multirow{2}{*}{DeepSeek-v3 685B} & \multirow{2}{*}{baseline} & zero-shot        & 42.01         & 64.37           & 68.88          \\
                              &                                      & one-shot  & 52.83          & 70.99          & 76.73          \\ \bottomrule
\end{tabular}
\end{small}

\caption{Evaluation of LLMs for \textbf{English-to-French medical translation}. After fine-tuning Llama-3 8B and NLLB-200 3.3 on the medium-sized dataset with fuzzy matches, they outperform all the evaluated models (\textbf{bold}). Llama-3.1 405B (one-shot) outperforms all the “baseline” models, while Llama-3 8B and Llama-3.1 8B (one-shot) are the best among baseline LLMs in the size range of 8B parameters. Nevertheless, for zero-shot translation, NLLB-200 3.3B is still the top-performer among all the evaluated baseline pre-trained models, except Llama-3.1 405B.}
\label{tab:mt-en-fr-medical}
\end{table*}

\begin{table*}[!ht]
\begin{small}

\centering
\begin{tabular}{@{}llllll@{}}
\toprule
\textbf{Model}                    & \textbf{Model Type}                  & \textbf{Context} & \textbf{BLEU}            & \textbf{chrF++}          & \textbf{COMET}           \\ \midrule
\multirow{8}{*}{NLLB-200 3.3B}    & \multirow{2}{*}{baseline}            & zero-shot        & 39.32                      & 64.98                      & 81.77                      \\
                                  &                                      & one-shot         & 39.79                      & 64.36                      & 79.04                      \\ \cmidrule(l){2-6} 
                                  & finetuned {[}s{]}                    & zero-shot        & 43.86                      & 67.52                      & 85.14                      \\ \cmidrule(l){2-6} 
                                  & finetuned {[}m{]}                    & zero-shot        & \uline{46.41} & \uline{69.13} & \uline{87.29} \\ \cmidrule(l){2-6} 
                                  & \multirow{2}{*}{finetuned {[}s+1{]}} & zero-shot        & 44.22                      & 67.82                      & 85.71                      \\
                                  &                                      & one-shot         & 43.74                      & 67.48                      & 85.34                      \\ \cmidrule(l){2-6} 
                                  & \multirow{2}{*}{finetuned {[}m+1{]}} & zero-shot        & 46.29                      & 69.08                      & 87.36                      \\
                                  &                                      & one-shot         & 45.90                      & 68.89                      & 87.15                      \\ \midrule
\multirow{4}{*}{Gemma 7B}         & \multirow{2}{*}{baseline}            & zero-shot        & 33.45                      & 57.91                      & 58.13                      \\
                                  &                                      & one-shot         & 40.39                      & 65.02                      & 80.59                      \\ \cmidrule(l){2-6} 
                                  & \multirow{2}{*}{finetuned {[}s+1{]}} & zero-shot        & 39.72                      & 64.10                      & 79.11                      \\
                                  &                                      & one-shot         & 40.28                      & 64.57                      & 79.98                      \\ \midrule
\multirow{6}{*}{Llama-3 8B}       & \multirow{2}{*}{baseline}            & zero-shot        & 38.68                      & 63.39                      & 77.68                      \\
                                  &                                      & one-shot         & 41.15                      & 65.49                      & 82.27                      \\ \cmidrule(l){2-6} 
                                  & \multirow{2}{*}{finetuned {[}s+1{]}} & zero-shot        & 42.10                      & 66.01                      & 82.48                      \\
                                  &                                      & one-shot         & 42.27                      & 66.07                      & 82.53                      \\ \cmidrule(l){2-6} 
                                  & \multirow{2}{*}{finetuned {[}m+1{]}} & zero-shot        & 43.63                      & 67.15                      & 84.22                      \\
                                  &                                      & one-shot         & 44.00                      & 67.37                      & 84.54                      \\ \midrule
\multirow{2}{*}{Llama-3.1 8B}     & \multirow{2}{*}{baseline}            & zero-shot        & 38.77                      & 63.51                      & 78.17                      \\
                                  &                                      & one-shot         & 41.44                      & 65.65                      & 82.47                      \\ \midrule
\multirow{2}{*}{Llama-3 70B}      & \multirow{2}{*}{baseline}            & zero-shot        & 43.86                      & 67.35                      & 84.05                      \\
                                  &                                      & one-shot         & 45.95                      & 68.82                      & 86.58                      \\ \midrule
\multirow{2}{*}{Llama-3.1 405B}   & \multirow{2}{*}{baseline}            & zero-shot        & 46.40                      & 69.08                      & 86.05                      \\
                                  &                                      & one-shot         & \textbf{47.76}             & \textbf{70.01}             & \textbf{88.07}             \\ \midrule
\multirow{4}{*}{Mistral 7B}       & \multirow{2}{*}{baseline}            & zero-shot        & 33.60                      & 60.08                      & 63.52                      \\
                                  &                                      & one-shot         & 37.64                      & 62.92                      & 78.05                      \\ \cmidrule(l){2-6} 
                                  & \multirow{2}{*}{finetuned {[}s+1{]}} & zero-shot        & 40.24                      & 64.77                      & 80.23                      \\
                                  &                                      & one-shot         & 40.62                      & 64.97                      & 80.98                      \\ \midrule
\multirow{2}{*}{Mixtral 8x7B}     & \multirow{2}{*}{baseline}            & zero-shot        & 39.58                      & 64.06                      & 77.46                      \\
                                  &                                      & one-shot         & 42.31                      & 66.33                      & 83.59                      \\ \midrule
\multirow{2}{*}{Qwen QwQ 32B}     & \multirow{2}{*}{baseline}            & zero-shot        & 42.43                      & 66.45                      & 83.62                      \\
                                  &                                      & one-shot         & 43.66                      & 67.31                      & 84.97                      \\ \midrule
\multirow{2}{*}{Qwen-2.5 72B}     & \multirow{2}{*}{baseline}            & zero-shot        & 44.11                      & 67.57                      & 85.14                      \\
                                  &                                      & one-shot         & 45.37                      & 68.40                      & 86.22                      \\ \midrule
\multirow{2}{*}{DeepSeek-v3 685B} & \multirow{2}{*}{baseline}            & zero-shot        & 45.84                      & 68.82                      & 86.08                      \\
                                  &                                      & one-shot         & 45.90                      & 68.59                      & 84.83                      \\ \bottomrule
\end{tabular}

\end{small}

\caption{Evaluation of LLMs for \textbf{English-to-Portuguese medical translation}. Llama-3.1 405B (one-shot) outperforms all the models (\textbf{bold}), while Llama-3 8B and Llama-3.1 8B (one-shot) are the best among LLMs in the size range of 8B parameters. After fine-tuning Llama-3 8B, it outperforms the decoder-only baseline models in the size range of 8B parameters a well as the fine-tuned Mistral, while Llama-3 70B and Llama-3.1 405B still outperform the fine-tuned Llama-3 8B. Nevertheless, for zero-shot translation, NLLB-200 3.3B is still the top-performer among all the evaluated pre-trained models in the size range of 8B parameters. After fine-tuning NLLB-200 3.3B with both the medium dataset (finetuned [m]), it outperforms all the models in the size range of 8B parameters (\uline{underlined}). Fine-tuning NLLB-200 3.3B on the sentence-level medium dataset (finetuned [m]) outperform fine-tuning it on the medium dataset concatenated with fuzzy matches (finetuned [m+1]).}
\label{tab:mt-en-pt-medical}
\end{table*}

\section{Evaluation and Discussion}
\label{sec:eval}

In this section, we discuss the evaluation results of our experiments. We experimented with both pure prompting pretrained LLMs for the task of translation, and fine-tuning these models. In both set of experiments, we evaluate two scenarios: (i) sentence-level translation (zero-shot), and (ii) retrieval-augmented translation through incorporating one in-context example (fuzzy match) into translation prompts (one shot). To evaluate our systems, we calculated BLEU \citep{Papineni2002-BLEU}, and chrF++ \citep{Popovic2017-chrF++}, as implemented in the sacreBLEU library \citep{Post2018-sacreBLEU}.\footnote{\url{https://github.com/mjpost/sacrebleu}} For semantic evaluation, we used COMET \cite{Rei2020-COMET}.\footnote{\url{https://github.com/Unbabel/COMET}}\textsuperscript{,}\footnote{In particular, we used the “wmt20-comet-da” model.} As COMET does not fully support Swahili, we used AfriCOMET-MTL (multitask learning) \citep{Wang2024-AfriCOMET} evaluation model.\footnote{\url{https://hf.co/masakhane/africomet-mtl}}

In this section, we explore the medical translation performance of a range of open-source models, including, Mistral, Mixtral, Llama, Gemma, Qwen, and DeepSeek, for both zero-shot and one-shot translation. For each language, we evaluate baselines (without fine-tuning). For models that show better performance, we also experimented with fine-tuning these LLMs, and hence we report the evaluation results of the fine-tuned versions too. Moreover, we evaluated NLLB-200 3.3B baseline and four fine-tuned variants (cf.~Section \ref{sec:experiments}).

\subsection{French and Portuguese}

In this section, we elaborate on the evaluation results for English-to-French (EN-FR) and English-to-Portuguese (EN-PT) medical translation (cf.~Table \ref{tab:mt-en-fr-medical} and Table~\ref{tab:mt-en-pt-medical}) using a range of open-source models.

\subsubsection{Baseline Evaluation} For both French and Portuguese, one-shot translation, i.e. prepending the prompt with a similar translation pair at inference time, outperforms zero-shot translation, using the baselines of all the covered autoregressive decoder-only LLMs. 

For zero-shot translation, NLLB-200 3.3B, the encoder-decoder MT model, outperforms all the baseline models in the size range of 8B parameters, while Llama-3 8B and Llama-3.1 one-shot translation outperforms the models in the same size range. Although Llama-3.1 models are promoted as being “multilingual” \citep{Dubey2024-Llama3}, Llama-3 8B and Llama 3.1 8B show comparable performance. 

When it comes to larger models, Mixtral 8x7B outperforms Llama 3 8B. Nevertheless, the gigantic Qwen-2.5 72B, Llama-3 70B, Llama-3.1 405B, and DeepSeek-v3 685B achieve the best performance among all the baseline models, especially with one-shot translation.

BioMistral \citep{Labrak2024-BioMistral} is an open-source LLM tailored for the biomedical domain, utilizing Mistral as its foundation model and further pre-trained on PubMed Central.
However, in our experiments, Mistral outperforms BioMistral at both zero-shot and one-shot EN-FR medical translation. Finally, Gemma 7B has the worst performance among all the models we evaluated for zero-shot translation. However, for one-shot translation, Gemma slightly outperforms Mistral.

\subsubsection{Fine-tuning Evaluation}

We fine-tuned Mistral 7B, Mixtral 7x8B, Llama-3 8B, and Gemma 7B. In this section, we elaborate on our evaluation results.

\paragraph{Decoder-only LLMs:} After fine-tuning Mistral, we observe considerable performance gains in both zero-shot and one-shot translation for both French and Portuguese (cf.~Table~\ref{tab:mt-en-fr-medical} and Table~\ref{tab:mt-en-pt-medical}). Interestingly, while fine-tuning Mixtral 8x7B enhances its performance for zero-shot translation, it degrades its one-shot translation performance. Fine-tuning both Mistral and Mixtral 8x7B, zero-shot translation of Mixtral 8x7B still outperforms \mbox{Mistral} 7B, while we do not observe the same level of improvement for one-shot translation. Fine-tuning Gemma 7B considerably improves zero-shot translation, while degrades one-shot EN-FR translation. Nevertheless, except for BLEU, Mistral still outperforms Gemma after fine-tuning both models for both zero-shot and one-shot translation based on all other metrics. The fine-tuned Llama-3 8B for EN-FR translation outperforms Llama-3 70B baseline (i.e. without fine-tuning) for zero-shot translation, while the evaluation scores for one-shot translation are relatively comparable. For EN-PT medical translation, the larger model Llama-3 70B still outperforms the fine-tuned Llama-3 8B model. 

\paragraph{NLLB-200 3.3:} Fine-tuning the task-oriented encoder-decoder MT model, NLLB-200 3.3B, reveals interesting findings. After fine-tuning on the medical dataset, NLLB-200 3.3B outperforms the fine-tuned Llama-3 8B at all the experiments in Portuguese (cf.~Table \ref{tab:mt-en-pt-medical}), while showing comparable results in French (cf.~Table \ref{tab:mt-en-fr-medical}). For French, NLLB-200 3.3B fine-tuned on the medium-sized dataset with concatenated fuzzy matches [m+1] (cf.~Section \ref{sec:data}) outperforms the gigantic Llama-3.1 405B and DeepSeek-v3 685B. For Portuguese, the best results are achieved when NLLB-200 3.3B is fine-tuned with a regular sentence-level medium-sized dataset [m], which outperforms both the baseline and fine-tuned Llama-3 8B models.

\newpage
It is worth mentioning that NLLB-200 3.3B was not originally pre-trained with fuzzy-match concatenation. Hence, one-shot translation might have quality issues in some translations, alleviated by fine-tuning. In other words, pre-training an encoder-decoder MT model with domain-specific fuzzy-match concatenation can be considered \citep{Bulte2019-fuzzy} if this feature is required later for fine-tuning and inference. Moreover, using quality estimation tools such as CometKiwi \citep{Rei2022-CometKiwi} or MetricX \citep{Juraska2024-MetricX} is recommended to select the best translation from zero-shot (sentence-level) and one-shot (fuzzy-match concatenated) translations. Nevertheless, it is important to evaluate the quality gains of fuzzy-match concatenation per se. Fuzzy-match concatenation diversifies and increases the data, which might explain improvements in zero-shot translation, but not one-shot translation. As in the case of Portuguese and Swahili (cf.~Section \ref{sec:swahili-eval}), conventional sentence-level fine-tuning can be sufficient. Other data augmentation techniques can be considered, including monolingual data back-translation \citep{Sennrich2016-BT,Edunov2018-BT} and domain-specific text generation \citep{Moslem2022-MT-LM}.

In a nutshell, when a small-to-medium-sized domain-specific dataset is available, fine-tuning the task-oriented NLLB-200 3.3B achieves comparable or better results than fine-tuning LLMs in the size range of 8B parameters such as Llama-3 8B and Mistral 7B, according to the chrF++ and COMET scores. In the absence of a domain-specific dataset, one-shot translation with decoder-only models, especially larger LLMs such as Mixtral 8x7B, Llama-3 70B and Llama-3.1 405B, outperforms all the baseline pre-trained models.

\subsection{Swahili}
\label{sec:swahili-eval}

\begin{table*}[!ht]
\centering
\begin{small}
\begin{tabular}{@{}lllllllll@{}}
\toprule
 &
   &
   &
  \multicolumn{3}{c}{\textbf{Medical}} &
  \multicolumn{3}{c}{\textbf{Generic}} \\ 
  \cmidrule(rl){4-6} \cmidrule(rl){7-9} \addlinespace[3pt]
  
\textbf{Model} &
  \textbf{Model Type} &
  \textbf{Context} &
  \textbf{BLEU} &
  \textbf{chrF++} &
  \textbf{COMET} &
  \textbf{BLEU} &
  \textbf{chrF++} &
  \textbf{COMET} \\ \midrule
\multirow{5}{*}{NLLB-200 3.3B} &
  \multirow{2}{*}{baseline} &
  zero-shot &
  44.58 &
  64.73 &
  84.11 &
  42.87 &
  61.59 &
  82.00 \\
 &
   &
  one-shot &
  43.66 &
  64.03 &
  83.92 &
  41.38 &
  59.99 &
  80.92 \\ \cmidrule(l){2-9} 
 &
  finetuned &
  zero-shot &
  45.80 &
  65.75 &
  83.98 &
  43.67 &
  62.06 &
  82.07 \\ \cmidrule(l){2-9} 
 & \smallskip
  \multirow{2}{*}{finetuned {[}+1{]}} &
  zero-shot &
  {\ul 47.11} &
  {\ul 66.70} &
  {\ul 84.29} &
  {\ul 44.25} &
  {\ul 62.49} &
  {\ul 82.35} \\
 &
   &
  one-shot &
  45.64 &
  65.85 &
  83.92 &
  43.31 &
  61.72 &
  82.04 \\ \midrule
\multirow{2}{*}{Gemma 7B} &
  \multirow{2}{*}{baseline} &
  zero-shot &
  35.20 &
  58.71 &
  81.70 &
  32.70 &
  54.87 &
  79.00 \\
 &
   &
  one-shot &
  38.98 &
  61.37 &
  83.02 &
  37.25 &
  58.01 &
  80.53 \\ \midrule
\multirow{4}{*}{Llama-3 8B} &
  \multirow{2}{*}{baseline} &
  zero-shot &
  29.19 &
  53.35 &
  79.77 &
  29.24 &
  52.97 &
  78.28 \\
 &
   &
  one-shot &
  35.91 &
  57.86 &
  81.60 &
  36.70 &
  56.84 &
  79.99 \\ \cmidrule(l){2-9} 
 &
  \multirow{2}{*}{finetuned {[}+1{]}} &
  zero-shot &
  38.33 &
  60.11 &
  82.16 &
  39.40 &
  58.77 &
  80.74 \\
 &
   &
  one-shot &
  39.47 &
  60.83 &
  82.24 &
  40.18 &
  59.04 &
  80.93 \\ \midrule
\multirow{2}{*}{Llama-3.1 8B} &
  \multirow{2}{*}{baseline} &
  zero-shot &
  28.19 &
  51.61 &
  78.06 &
  28.62 &
  51.79 &
  76.90 \\
 &
   &
  one-shot &
  33.92 &
  56.11 &
  80.54 &
  35.09 &
  55.49 &
  78.92 \\ \midrule
\multirow{2}{*}{Llama-3 70B} &
  \multirow{2}{*}{baseline} &
  zero-shot &
  38.31 &
  61.14 &
  82.27 &
  36.75 &
  59.02 &
  81.05 \\
 &
   &
  one-shot &
  44.06 &
  64.31 &
  83.36 &
  42.27 &
  61.52 &
  81.84 \\ \midrule
\multirow{2}{*}{Llama-3.1 405B} &
  \multirow{2}{*}{baseline} &
  zero-shot &
  47.76 &
  66.57 &
  83.12 &
  42.21 &
  61.87 &
  81.70 \\
 &
   &
  one-shot &
  \textbf{52.00} &
  \textbf{68.79} &
  \textbf{83.23} &
  \textbf{45.61} &
  \textbf{63.46} &
  \textbf{82.10} \\ \midrule
\multirow{4}{*}{Mistral 7B} &
  \multirow{2}{*}{baseline} &
  zero-shot &
  11.94 &
  32.67 &
  59.12 &
  17.04 &
  36.69 &
  59.83 \\
 &
   &
  one-shot &
  16.74 &
  36.83 &
  65.26 &
  21.44 &
  40.16 &
  63.27 \\ \cmidrule(l){2-9} 
 &
  \multirow{2}{*}{finetuned {[}+1{]}} &
  zero-shot &
  31.62 &
  52.79 &
  77.67 &
  34.64 &
  53.02 &
  76.04 \\
 &
   &
  one-shot &
  32.50 &
  53.73 &
  78.59 &
  35.08 &
  53.36 &
  76.24 \\ \midrule
\multirow{2}{*}{Mixtral 8x7B} &
  \multirow{2}{*}{baseline} &
  zero-shot &
  19.32 &
  41.50 &
  69.85 &
  24.87 &
  45.14 &
  70.30 \\
 &
   &
  one-shot &
  24.09 &
  45.04 &
  73.23 &
  28.53 &
  47.98 &
  72.46 \\ \midrule
\multirow{2}{*}{Qwen QwQ 32B} &
  \multirow{2}{*}{baseline} &
  zero-shot &
  25.00 &
  48.55 &
  74.96 &
  26.07 &
  48.56 &
  72.61 \\
 &
   &
  one-shot &
  30.41 &
  52.49 &
  77.25 &
  31.05 &
  51.34 &
  74.43 \\ \midrule
\multirow{2}{*}{Qwen-2.5 72B} &
  \multirow{2}{*}{baseline} &
  zero-shot &
  28.34 &
  52.01 &
  76.40 &
  29.92 &
  51.75 &
  74.85 \\
 &
   &
  one-shot &
  35.09 &
  56.03 &
  78.67 &
  35.32 &
  54.55 &
  76.82 \\ \bottomrule
\end{tabular}
\end{small}
\caption{Evaluation of LLMs for \textbf{Swahili-to-English medical and generic translation}. Llama-3.1 405B (one-shot) is the top-performing model (\textbf{bold}). Nevertheless, the encoder-decoder MT model, NLLB-200 3.3B (both baseline and fine-tuned) is the top-performer among the models in the size range of 8B parameters (\uline{underlined}).}
\label{tab:mt-sw-en}
\end{table*}

Based on the evaluation results for both the generic and medical domain, the top-performing model for Swahili-to-English (SW-EN) and English-to-Swahili translation continues to be NLLB-200 3.3B, a Transformer-based encoder-decoder model specifically developed for MT tasks.

As illustrated by Table \ref{tab:mt-sw-en}, in both the medical and generic domains of Swahili translation, Gemma and Llama-3 8B (especially after fine-tuning) are the top performers among autoregressive decoder-only LLMs in the size range of 8B parameters. However, NLLB-200 3.3B outperforms these models. Nevertheless, Llama-3 70B evaluation scores are relatively comparable to those of NLLB-200. 

\begin{table*}[!ht]
\centering
\begin{small}

\begin{tabular}{@{}lllllllll@{}}
\toprule
 &
   &
   &
  \multicolumn{3}{c}{\textbf{Medical}} &
  \multicolumn{3}{c}{\textbf{Generic}} \\ 
  \cmidrule(rl){4-6} \cmidrule(rl){7-9} \addlinespace[3pt]

\textbf{Model} &
  \textbf{Model Type} &
  \textbf{Context} &
  \textbf{BLEU} &
  \textbf{chrF++} &
  \textbf{COMET} &
  \textbf{BLEU} &
  \textbf{chrF++} &
  \textbf{COMET} \\ \midrule
\multirow{5}{*}{NLLB-200 3.3B} &
  \multirow{2}{*}{baseline} &
  zero-shot &
  32.61 &
  58.24 &
  82.85 &
  35.57 &
  57.55 &
  83.90 \\
 &
   &
  one-shot &
  34.06 &
  59.12 &
  82.37 &
  33.79 &
  55.93 &
  83.23 \\ \cmidrule(l){2-9} 
 &
  finetuned &
  zero-shot &
  36.21 &
  61.35 &
  83.92 &
  38.98 &
  60.21 &
  85.05 \\ \cmidrule(l){2-9} 
 &
  \multirow{2}{*}{finetuned {[}+1{]}} &
  zero-shot &
  37.31 &
  62.37 &
  84.18 &
  \textbf{39.58} &
  \textbf{60.78} &
  \textbf{85.45} \\
 &
   &
  one-shot &
  \textbf{37.52} &
  \textbf{62.29} &
  \textbf{83.93} &
  39.09 &
  60.34 &
  85.03 \\ \midrule
\multirow{2}{*}{Gemma 7B} &
  \multirow{2}{*}{baseline} &
  zero-shot &
  11.66 &
  29.87 &
  60.62 &
  \hspace{0.5em}8.32 &
  21.51 &
  54.88 \\
 &
   &
  one-shot &
  21.37 &
  47.15 &
  71.57 &
  22.51 &
  45.37 &
  71.08 \\ \midrule
\multirow{4}{*}{Llama-3 8B} &
  \multirow{2}{*}{baseline} &
  zero-shot &
  10.10 &
  33.82 &
  60.95 &
  13.25 &
  34.34 &
  61.68 \\
 &
   &
  one-shot &
  16.08 &
  42.38 &
  67.03 &
  20.03 &
  43.80 &
  69.56 \\ \cmidrule(l){2-9} 
 &
  \multirow{2}{*}{finetuned {[}+1{]}} &
  zero-shot &
  21.50 &
  47.60 &
  69.94 &
  26.47 &
  49.24 &
  73.74 \\
 &
   &
  one-shot &
  23.58 &
  48.99 &
  71.69 &
  28.30 &
  50.15 &
  74.40 \\ \midrule
\multirow{2}{*}{Llama-3.1 8B} &
  \multirow{2}{*}{baseline} &
  zero-shot &
  \hspace{0.5em}6.33 &
  28.06 &
  51.69 &
  10.17 &
  30.44 &
  55.64 \\
 &
   &
  one-shot &
  11.64 &
  36.91 &
  58.98 &
  16.10 &
  39.54 &
  62.61 \\ \midrule
\multirow{2}{*}{Llama-3 70B} &
  \multirow{2}{*}{baseline} &
  zero-shot &
  24.34 &
  49.49 &
  77.97 &
  25.77 &
  48.20 &
  77.67 \\
 &
   &
  one-shot &
  28.06 &
  53.67 &
  79.87 &
  31.58 &
  53.82 &
  81.71 \\ \midrule
\multirow{2}{*}{Llama-3.1 405B} &
  \multirow{2}{*}{baseline} &
  zero-shot &
  29.65 &
  55.19 &
  80.21 &
  32.53 &
  53.98 &
  80.82 \\
 &
   &
  one-shot &
  31.65 &
  57.27 &
  81.41 &
  35.07 &
  57.06 &
  83.57 \\ \midrule
\multirow{4}{*}{Mistral 7B} &
  \multirow{2}{*}{baseline} &
  zero-shot &
  \hspace{0.5em}0.64 &
  11.02 &
  25.79 &
  \hspace{0.5em}1.82 &
  13.15 &
  30.15 \\
 &
   &
  one-shot &
  \hspace{0.5em}1.20 &
  14.67 &
  22.72 &
  \hspace{0.5em}2.70 &
  17.02 &
  27.88 \\ \cmidrule(l){2-9} 
 &
  \multirow{2}{*}{finetuned {[}+1{]}} &
  zero-shot &
  12.88 &
  37.76 &
  46.65 &
  19.28 &
  42.12 &
  53.35 \\
 &
   &
  one-shot &
  14.83 &
  39.81 &
  49.91 &
  21.04 &
  43.46 &
  55.01 \\ \midrule
\multirow{2}{*}{Mixtral 8x7B} &
  \multirow{2}{*}{baseline} &
  zero-shot &
  \hspace{0.5em}2.17 &
  18.34 &
  31.47 &
  \hspace{0.5em}5.35 &
  20.89 &
  40.93 \\
 &
   &
  one-shot &
  \hspace{0.5em}4.00 &
  24.88 &
  36.66 &
  \hspace{0.5em}7.73 &
  28.52 &
  43.73 \\ \midrule
\multirow{2}{*}{Qwen QwQ 32B} &
  \multirow{2}{*}{baseline} &
  zero-shot &
  5.17 &
  27.18 &
  38.72 &
  7.73 &
  28.90 &
  42.51 \\
 &
   &
  one-shot &
  9.38 &
  33.48 &
  44.50 &
  13.19 &
  35.49 &
  48.23 \\ \midrule
\multirow{2}{*}{Qwen-2.5 72B} &
  \multirow{2}{*}{baseline} &
  zero-shot &
  7.57 &
  30.88 &
  45.96 &
  10.40 &
  32.30 &
  50.88 \\
 &
   &
  one-shot &
  10.52 &
  35.95 &
  50.54 &
  15.24 &
  38.48 &
  54.82 \\ \bottomrule
\end{tabular}
\end{small}
\caption{Evaluation of LLMs for the \textbf{English-to-Swahili medical and generic translation}. NLLB-200 3.3B (both baseline and fine-tuned) is the top-performer among all the models, even those with larger sizes.}
\label{tab:mt-en-sw}
\end{table*}

As expected, the evaluation results for SW-EN translation with LLMs are superior to those of EN-SW translation due to the limited support for lower-resource languages such as Swahili. Similar to SW-EN translation, the top model for EN-SW is still the NLLB-200 3.3. For both the medical domain and the generic domain (cf.~Table~\ref{tab:mt-en-sw}) of EN-SW translation, all the autoregressive decoder-only models in our experiments demonstrate a lower translation quality. Even after fine-tuning Llama-3 8B, its evaluation scores are not on par with those of NLLB-200 3.3.

Although Llama 3.1 8B is promoted as a multilingual model \cite{Dubey2024-Llama3}, Llama 3 8B outperforms it for EN-SW translation. Comparing model sizes,  Llama 3 70B outperforms the smaller version of the model Llama 3 8B for both zero-shot and one-shot EN-SW translation.

\paragraph{Fine-tuning:} We fine-tuned Mistral~7B and Llama-3~8B for both generic and medical SW-EN and EN-SW translation. While fine-tuning has improved the evaluation scores compared to the baselines, fine-tuning NLLB-200 3.3B outperforms even Llama-3~70B for SW-EN translation and both Llama-3 70B and Llama-3.1~405B for EN-SW translation.

\section{Conclusions}

After a few years of research on leveraging open-source LLMs for direct translation among other tasks, several businesses have been investigating the feasibility of integrating such general-purpose models into their specialized translation workflows. Among the main advantages of employing the in-context learning feature of LLMs is maintaining consistency, domain terminology and style throughout the entire document \cite{Dettmers2023-QLoRA,Wang2023-DocumentMT}. This can be achieved by feeding an LLM with similar translation pairs, preapproved terms, or other useful data at inference time, hoping that the LLM's output will be influenced by the provided linguistic data. Moreover, several businesses and organizations have resorted to open-source LLMs for data privacy and security reasons. Nevertheless, deploying open-source LLMs into production poses efficiency challenges, including infrastructure requirements, latency at translation time, and inconsistent quality. This raises the question of whether conventional encoder-decoder MT models remain relevant, particularly for domain-specific translation.

As elaborated in Section~\ref{sec:experiments}, we evaluated a range of LLMs in the medical domain. Moreover, we fine-tuned selected models for zero-shot and one-shot translation. Our experiments covered both the multilingual encoder-decoder MT model, NLLB-200~3.3, and several autoregressive decoder-only LLMs, such as Gemma, Llama, and Mistral. Our retrieval-augmented translation used one-shot in-context examples based on semantic similarly with the new source (fuzzy match), which achieves better results than random selection \citep{Agrawal2023-SelectionMT,Moslem2023-AdaptiveMT}. In the future, we would like to compare retrieval-augmented context (as in this work) with direct context from previous and next sentences in the document.

For French and Portuguese, some baselines of decoder-only LLMs, such as Llama 3 8B and Mixtral 8x7B, outperform NLLB-200 3.3B. However, \mbox{after} fine-tuning NLLB-200 3.3B, the performance gap narrows considerably, and NLLB performs on par with or even surpasses the decoder-only LLMs, especially models in the size range of 8B parameters. For Swahili, only the massive Llama 3.1 405B model achieves performance comparable to NLLB-200 3.3B, while all other evaluated decoder-only LLMs fall short.

Our results demonstrate the diminishing returns of using decoder-only LLMs in the size range of 8B parameters for domain-specific translation, especially when a small-to-medium-sized dataset is available for fine-tuning task-oriented translation models like NLLB-200 3.3. Larger LLMs, e.g. Llama-3 70B, Llama 3.1 405B, or Mixtral 8x7B, can be considered for some tasks such as real-time adaptive translation or automatic post-editing, as they offer better quality than their 8B-parameter variants. However, their massive sizes make them less efficient for real-time scenarios in production, and more suitable for offline data augmentation, or knowledge distillation into more efficient models. This presents a valuable opportunity for businesses to develop in-house medium-sized LMs, custom-built for their linguistic needs, especially for domain-specific and low-resource scenarios. In the meantime, task-oriented encoder-decoder MT models remain a core component in high-quality domain-specific translation workflows.

\section{Computing Considerations}

We are making the code we used for both training and inference publicly available\footnote{\url{https://github.com/ymoslem/Adaptive-MT-LLM-Fine-tuning}} to help the community reduce experimentation runtime and cost. Moreover, we would like to share some facts and recommendations:
\begin{itemize}
    \setlength{\topsep}{0pt}
    \setlength{\itemsep}{0pt}
    \item Both inference and fine-tuning of LLMs in the range of 8B parameters was conducted on 1x A100 40GB GPU. For efficient fine-tuning, we used QLoRA (cf.~Section \ref{sec:experiments}).
    \item For inference of \textit{Llama-3~70B} and \textit{Qwen-2.5 72B} (without quantization), we used 2x H100 GPUs, with the vLLM inference engine. 
    \item For inference of \textit{Llama-3.1 405B} with AWQ quantization, we used 4x H100 GPUs, while for \textit{DeepSeek-V3 685B} with AWQ quantization, we used 4x H200 GPUs (alternatively, 8x H100 GPUs are required), with the vLLM inference engine.
    \item Using \textit{Qwen-2.5 72B} to translate the English-to-Portuguese test set with one-shot prompts took approximately 20 minutes, while using \textit{DeepSeek-V3 685B} to translate the same dataset took approximately 2 hours (in addition to the time of downloading and loading the model).
    \item We recommend using CTranslate2 and vLLM as efficient inference engines (cf.~Table \ref{tab:inference-comparison}).
    \item While we only used quantization with Llama-3.1 405B and DeepSeek-V3 685B due to their massive sizes, we recommend considering quantization at inference time, as it does not significantly affect the translation quality (cf.~Appendix \ref{sec:appendix-inference}).
\end{itemize}

\section{Limitations}

Since all the data used in this work has been publicly available for years, there is a potential for data contamination. Many open-source LLMs do not disclose the specific datasets used for training, so we cannot be certain whether our test datasets were part of the training data for any of these models. However, we conducted multiple experiments for each model, including sentence-level translation (zero-shot), retrieval-augmented translation with fuzzy matches (one-shot), and fine-tuning of some models. As such, we consider our results a reliable indication of the translation quality that can be expected from these LLMs in similar setups. It is important to note that, as this paper demonstrates, results may vary across different languages or domains.

\bibliography{paperpile}

\clearpage
\appendix 

\newpage

\section{Appendix: Data}
\label{sec:appendix-data}

As explained in Section \ref{sec:data}, we collected public datasets in the medical domain, filtered them, and sampled portions for our experiments. This section further clarifies the data distribution and preparation outcomes in terms of (i) filtering (cf.~Table \ref{tab:data-french}, \ref{tab:data-portuguese} \& \ref{tab:data-swahili}), and (ii) sampling (cf.~Table \ref{tab:data-sampled}).

\subsection{Filtering}

The filtering step removes duplicates, too long segments, and semantically mismatching segments. The scripts of our data preparation are publicly available.\footnote{\url{https://github.com/ymoslem/MT-Preparation/}}

\begin{table}[ht]
\centering
\resizebox{\columnwidth}{!}{%
\begin{tabular}{@{}lrrr@{}}
\toprule
\textbf{Dataset}             & \textbf{Raw}        & \textbf{Filtered} \\ \midrule
ELRC\_2922                  & 4,365      & 4,022  \\
ELRC\_2923                  & 502        & 470  \\
ELRC\_3382                  & 6,593      & 6,537  \\
ELRC-antibiotic             & 1,072      & 1,009  \\
ELRC-EUROPARL\_covid        & 789        & 747  \\
ELRC-presscorner\_covid     & 11,361     & 11,265  \\
ELRC-wikipedia\_health      & 4,366      & 4,045  \\
EMEA                        & 1,092,568  & 292,098  \\
Tico-19                     & 3,071      & 2,768  \\ \midrule
Total                       & 1,124,687  & 322,961  \\ \midrule
\textbf{Global Filtering} & \multicolumn{1}{l}{} & 303,948      & \multicolumn{1}{l}{} \\ \bottomrule
\end{tabular}
}

\caption{Medical datasets for French}
\label{tab:data-french}
\end{table}
\begin{table}[ht]
\centering
\resizebox{\columnwidth}{!}{%
\begin{tabular}{@{}lrrr@{}}
\toprule

\textbf{Dataset }        & \textbf{Raw}        & \textbf{Filtered}       \\ \midrule
ELRC\_2922                  & 7,292        & 7,129  \\
ELRC\_2923                  & 560          & 519  \\
ELRC\_3382                  & 3,513        & 3,465  \\
ELRC-antibiotic             & 839          & 775  \\
ELRC-EUROPARL\_covid        & 864          & 803  \\
ELRC-presscorner\_covid     & 6,319        & 6,239  \\
ELRC-wikipedia\_health      & 7,293        & 7,153  \\
EMEA                        & 1,082,144    & 290,782  \\
SciELO                      & 3,084,830    & 2,716,409   \\
Tico-19                     & 3,071        & 3,067  \\ \midrule
Total                       & 4,196,725    & 3,036,341 \\ \midrule
\textbf{Global Filtering} & \multicolumn{1}{l}{} & 2,905,418      & \multicolumn{1}{l}{} \\ \bottomrule
\end{tabular}
}

\caption{Medical datasets for Portuguese}
\label{tab:data-portuguese}
\end{table}
\begin{table}[ht]
\centering
\resizebox{\columnwidth}{!}{%
\begin{tabular}{@{}llrrr@{}}
\toprule

\textbf{Domain} & \textbf{Dataset} & \textbf{Raw} & \textbf{Filtered} \\ \midrule
\multirow{2}{*}{Medical} & ELRC-wikipedia\_health & 608 &  593 \\
& Tico-19   & 3,071 &  3,056 \\ \cmidrule(l){2-5}
& Total     & 3,679 &  3,649 \\ \cmidrule(l){2-5}
& \textbf{Global Filtering} & \multicolumn{1}{l}{} & 3,514 & \multicolumn{1}{l}{} \\
\midrule
\multirow{2}{*}{Generic} & GlobalVoices & 32,307 &  29,623 \\
& Wikimedia & 16,289 &  13,829 \\ \cmidrule(l){2-5}
& Total     & 48,596 &  43,452 \\ \cmidrule(l){2-5}
& \textbf{Global Filtering} & \multicolumn{1}{l}{} & 39,409 & \multicolumn{1}{l}{} \\
\bottomrule
\end{tabular}
}

\caption{Medical and generic datasets for Swahili}
\label{tab:data-swahili}
\end{table}



\subsection{Sampling and Splitting}

This step samples portions of the datasets, and splits them as follows: (i) training dataset, (ii) test dataset (iii) train/test context datasets, i.e. two unique translation memories used to retrieve fuzzy matches that are used in one-shot prompts.

\begin{table}[ht]
\centering
\resizebox{\columnwidth}{!}{%

\begin{tabular}{@{}lllllll@{}}
\toprule
\multicolumn{1}{c}{\multirow{2}{*}{\textbf{Lang}}} &
  \multirow{2}{*}{\textbf{Domain}} &
  \multicolumn{3}{c}{\textbf{Train}} &
  \multicolumn{2}{c}{\textbf{Test}} \\ \cmidrule(lr){3-5} \cmidrule(lr){6-7}
\multicolumn{1}{c}{} &
   &
  \multicolumn{1}{c}{\textbf{small}} &
  \multicolumn{1}{c}{\textbf{medium}} &
  \multicolumn{1}{c}{\textbf{context}} &
  \multicolumn{1}{c}{\textbf{test}} &
  \multicolumn{1}{c}{\textbf{context}} \\ \midrule
\textbf{French}     & Medical & 10,000 & 100,000 & 50,000 & 10,000 & 50,000 \\ \midrule
\textbf{Portuguese} & Medical & 10,000 & 100,000 & 50,000 & 10,000 & 50,000 \\ \midrule
\multirow{2}{*}{\textbf{Swahili}} &
  Generic &
  \multirow{2}{*}{\begin{tabular}[c]{@{}l@{}}9,000\\ + 1,000\end{tabular}} &
  \multirow{2}{*}{n/a} &
  \multirow{2}{*}{\begin{tabular}[c]{@{}l@{}}10,415\\ + 514\end{tabular}} &
  10,000 &
  10,000 \\
                    & Medical &        &         &        & 1,000  & 1,000  \\ \bottomrule
\end{tabular}

}

\caption{Statistics of data that was actually used for training, testing, and context datasets.}
\label{tab:data-sampled}
\end{table}

\subsection{LLM Prompts}
\label{sec:prompts}

\begin{footnotesize}

\textbf{Zero-shot prompt example:}

\begin{itemize}
    \setlength{\topsep}{-2pt}
    \setlength{\itemsep}{-2pt}
    \item[] English: <source\_sentence>
    \item[] Swahili:
\end{itemize}

\noindent\textbf{One-shot prompt example:}

\begin{itemize}
    \setlength{\topsep}{-2pt}
    \setlength{\itemsep}{-2pt}
    \item[] English: <source\_fuzzy\_sentence>
    \item[] Swahili: <target\_fuzzy\_sentence>
    \item[] English: <source\_sentence>
    \item[] Swahili:
\end{itemize}

\end{footnotesize}

\section{Appendix: Training}
\label{sec:appendix-training}

In this appendix, we briefly summarize the results of extra experiments we conducted. The experiments include various embedding models used to create fuzzy matches, the impact of learning rate choice, dataset packing, and comparing inference engines, CTranslate2 and vLLM.
Moreover, we compare quantization approaches in Section \ref{sec:quantization}. We use the same evaluation metics BLEU, chrF++, TER, and COMET. For Swahili, the COMET column shows AfriCOMET-MTL scores.

\subsection{Embedding Models}

To create embeddings used later to generate the \mbox{retrieval} index (cf.~Section~\ref{sec:experiments}), we mainly used \mbox{Microsoft} “\textit{Multilingual-MiniLM-L12-H384}” throughout our paper, as it is a multilingual model. Furthermore, we experimented with “\textit{all-MiniLM-L6-v2}” and “\textit{all-mpnet-base-v2}”, which are an efficient embedding models for English. As we retrieve translation pairs based on the source text, we experimented with English-to-Portuguese and English-to-Swahili language directions. Table~\ref{tab:embedding-models} shows evaluation of the Mistral~7B model, both the baseline and fine-tuned version, using \mbox{CTranslate2} with int8 quantization. For English-to-Portuguese, the results were comparable, while for English-to-Swahili the results achieved by using Microsoft “\textit{Multilingual-MiniLM-L12-H384}” were generally better than the two other embedding models, so we used it for all our experiments.

\begin{table}[ht]
\centering
\resizebox{\columnwidth}{!}{
\begin{tabular}{@{}lllllcccc@{}}
\toprule
\textbf{Lang} & \textbf{Model} & \textbf{Domain} & \textbf{Context} & \textbf{Embedding} & \textbf{BLEU ↑} & \textbf{ChrF++ ↑} & \textbf{TER ↓} & \textbf{COMET ↑} \\
\midrule
\multirow{12}{*}{\textbf{EN-PT}} & \multirow{6}{*}{baseline} & \multirow{6}{*}{Medical} & \multirow{1}{*}{zero-shot} & N/A & 32.70 & 59.81 & 56.39 & 59.92 \\
\cmidrule(l){4-9}
& & & \multirow{3}{*}{one-shot} & MiniLM & 37.43 & \textbf{62.89} & 49.44 & 78.01 \\
& & & & MPNet & \textbf{37.49} & 62.82 & \textbf{49.33} & \textbf{78.25} \\
& & & & Multi-MiniLM & 37.41 & 62.81 & 49.47 & 77.91 \\
\cmidrule(l){2-9}
& \multirow{6}{*}{finetuned} & \multirow{6}{*}{Medical} & \multirow{3}{*}{zero-shot} & MiniLM & 39.89 & 64.39 & 47.16 & 79.77 \\
& & & & MPNet & 39.44 & 64.05 & 47.95 & 79.22 \\
& & & & Multi-MiniLM & \textbf{40.05} & \textbf{64.68} & \textbf{47.08} & \textbf{80.12} \\
\cmidrule(l){4-9}
& & & \multirow{3}{*}{one-shot} & MiniLM & \textbf{40.58} & \textbf{64.92} & \textbf{46.49} & \textbf{81.18} \\
& & & & MPNet & 40.20 & 64.57 & 47.32 & 80.59 \\
& & & & Multi-MiniLM & 40.39 & 64.86 & 47.11 & 80.61 \\
\midrule
\multirow{16}{*}{\textbf{EN-SW}} & \multirow{8}{*}{baseline} & \multirow{4}{*}{Medical} & \multirow{1}{*}{zero-shot} & N/A & 0.49 & 10.22 & 290.16 & 25.34 \\
\cmidrule(l){4-9}
& & & \multirow{2}{*}{one-shot} & MiniLM & \textbf{1.48} & \textbf{14.67} & 261.35 & \textbf{24.09} \\
& & & & Multi-MiniLM & 0.97 & 13.18 & \textbf{283.04} & 21.87 \\
\cmidrule(l){3-9}
& & \multirow{4}{*}{Generic} & \multirow{1}{*}{zero-shot} & N/A & 1.30 & 11.50 & 278.97 & 28.16 \\
\cmidrule(l){4-9}
& & & \multirow{2}{*}{one-shot} & MiniLM & \textbf{2.26} & \textbf{15.17} & \textbf{278.88} & \textbf{27.15} \\
& & & & Multi-MiniLM & 1.99 & 14.45 & 294.13 & 25.65 \\
\cmidrule(l){2-9}
& \multirow{8}{*}{finetuned} & \multirow{4}{*}{Medical} & \multirow{2}{*}{zero-shot} & MiniLM & 8.83 & 33.11 & 140.96 & 42.44 \\
& & & & Multi-MiniLM & \textbf{13.38} & \textbf{38.06} & \textbf{96.06} & \textbf{46.49} \\
\cmidrule(l){4-9}
& & & \multirow{2}{*}{one-shot} & MiniLM & 11.10 & 36.81 & 128.07 & 47.81 \\
& & & & Multi-MiniLM & \textbf{15.67} & \textbf{40.67} & \textbf{84.47} & \textbf{49.57} \\
\cmidrule(l){3-9}
& & \multirow{4}{*}{Generic} & \multirow{2}{*}{zero-shot} & MiniLM & 12.21 & 34.95 & 147.31 & 47.47 \\
& & & & Multi-MiniLM & \textbf{19.12} & \textbf{41.71} & \textbf{94.07} & \textbf{52.85} \\
\cmidrule(l){4-9}
& & & \multirow{2}{*}{one-shot} & MiniLM & 13.97 & 37.32 & 133.67 & 50.51 \\
& & & & Multi-MiniLM & \textbf{21.12} & \textbf{43.16} & \textbf{85.21} & \textbf{54.37} \\
\bottomrule
\end{tabular}
}
\caption{Comparison of embedding models for English-to-Portuguese (EN-PT) and English-to-Swahili (EN-SW) translation, showing baseline and fine-tuned performance on medical and generic test datasets. The table illustrates the performance of all-MiniLM-L6-v2 (MiniLM), Microsoft's Multilingual-MiniLM-L12-H384 (Multi-MiniLM), and All-MPNet-Base-V2 (MPNet). The embedding models are used to encode fuzzy matches before creating the retrieval index; that is why this step applies to one-shot translation only, when using pretrained models. Still, this step can affect the quality of fine-tuned models, as the data includes both zero-shot and one-shot prompts.}
\label{tab:embedding-models}
\end{table}

\subsection{Learning Rate}
Before conducting the main experiments, we tested a range of learning rate values to identify the optimal value.

\subsubsection{Mistral}
The learning rates we explored while fine-tuning Mistral 7B include 3e-4, 1e-3 and 2e-3. The results are presented in Table \ref{tab:lr-mistral}. Overall, the learning rate of 1e-3 produced the highest performance. For the sake of consistency, we then used the learning rate 1e-3 for fine-tuning all the decoder-only LLMs.

\subsubsection{NLLB}

Similarly, when fine-tuning NLLB-200 3.3B, we tried a few learning rates. Unlike Mistral, the learning rate 1e-3 did not achieve the best result. However, the learning rate 5e-5 revealed better evaluation scores, as illustrated by Table \ref{tab:lr-nllb}. Hence, we used it when fine-tuning NLLB-200 3.3B for all the language pairs.

\begin{table}[ht]
\centering
\resizebox{\columnwidth}{!}{%

\begin{tabular}{@{}llllcccc@{}}
\toprule
\textbf{Lang} & \textbf{Domain} & \textbf{Context} & \textbf{LR} & \textbf{BLEU ↑} & \textbf{ChrF++ ↑} & \textbf{TER ↓} & \textbf{COMET ↑} \\
\midrule
\multirow{4}{*}{\textbf{EN-FR}} & \multirow{4}{*}{medical} & \multirow{2}{*}{zero-shot} & 1e-3 & 39.33 & \textbf{64.57} & 51.18 & \textbf{71.89} \\
& & & 2e-3 & \textbf{41.73} & 64.45 & \textbf{50.15} & 71.47 \\
\cmidrule(l){3-8}
& & \multirow{2}{*}{one-shot} & 1e-3 & 45.31 & 68.57 & 45.81 & 74.04 \\
& & & 2e-3 & \textbf{48.15} & \textbf{68.94} & \textbf{44.55} & \textbf{76.21} \\
\midrule
\multirow{8}{*}{\textbf{EN-PT}} & \multirow{8}{*}{medical} & \multirow{4}{*}{zero-shot} & 1e-3 & \textbf{40.05} & \textbf{64.68} & 47.08 & \textbf{80.12} \\
& & & 2e-3 & 38.63 & 63.33 & 49.55 & 77.30 \\
& & & 1e-4 & 38.90 & 63.80 & 48.08 & 78.90 \\
& & & 3e-4 & 39.82 & 64.46 & \textbf{47.14} & 80.06 \\
\cmidrule(l){3-8}
& & \multirow{4}{*}{one-shot} & 1e-3 & 40.39 & \textbf{64.86} & 47.11 & 80.61 \\
& & & 2e-3 & 39.00 & 63.63 & 48.74 & 78.84 \\
& & & 1e-4 & 39.43 & 64.11 & 47.70 & 79.89 \\
& & & 3e-4 & \textbf{40.30} & 64.78 & \textbf{46.86} & \textbf{80.91} \\
\midrule
\multirow{12}{*}{\textbf{EN-SW}} & \multirow{6}{*}{medical} & \multirow{3}{*}{zero-shot} & 1e-3 & \textbf{13.38} & \textbf{38.06} & \textbf{96.06} & \textbf{46.49} \\
& & & 2e-3 & 11.61 & 36.35 & 113.01 & 45.55 \\
& & & 3e-4 & 4.96 & 25.29 & 162.21 & 31.13 \\
\cmidrule(l){3-8}
& & \multirow{3}{*}{one-shot} & 1e-3 & \textbf{15.67} & \textbf{40.67} & \textbf{84.47} & \textbf{49.57} \\
& & & 2e-3 & 14.46 & 39.46 & 96.04 & 48.70 \\
& & & 3e-4 & 6.65 & 28.24 & 147.95 & 35.71 \\
\cmidrule(l){2-8}
& \multirow{6}{*}{generic} & \multirow{3}{*}{zero-shot} & 1e-3 & 19.12 & \textbf{41.71} & \textbf{94.07} & \textbf{52.85} \\
& & & 2e-3 & \textbf{19.70} & 41.55 & 91.85 & 53.63 \\
& & & 3e-4 & 11.21 & 32.51 & 136.13 & 40.66 \\
\cmidrule(l){3-8}
& & \multirow{3}{*}{one-shot} & 1e-3 & 21.12 & \textbf{43.16} & \textbf{85.21} & \textbf{54.37} \\
& & & 2e-3 & \textbf{21.66} & 42.88 & 83.99 & 54.98 \\
& & & 3e-4 & 12.53 & 34.17 & 129.39 & 42.99 \\
\midrule
\multirow{8}{*}{\textbf{SW-EN}} & \multirow{4}{*}{medical} & \multirow{2}{*}{zero-shot} & 1e-3 & \textbf{31.81} & \textbf{52.87} & \textbf{58.53} & \textbf{77.38} \\
& & & 2e-3 & 27.03 & 47.81 & 63.38 & 73.69 \\
\cmidrule(l){3-8}
& & \multirow{2}{*}{one-shot} & 1e-3 & \textbf{32.25} & \textbf{53.50} & \textbf{57.52} & \textbf{78.46} \\
& & & 2e-3 & 27.44 & 48.21 & 63.37 & 74.89 \\
\cmidrule(l){2-8}
& \multirow{4}{*}{generic} & \multirow{2}{*}{zero-shot} & 1e-3 & \textbf{34.49} & \textbf{52.93} & \textbf{58.24} & \textbf{75.87} \\
& & & 2e-3 & 31.97 & 50.17 & 61.07 & 73.84 \\
\cmidrule(l){3-8}
& & \multirow{2}{*}{one-shot} & 1e-3 & \textbf{34.97} & \textbf{53.28} & \textbf{57.21} & \textbf{76.06} \\
& & & 2e-3 & 32.29 & 50.45 & 60.71 & 74.22 \\
\bottomrule
\end{tabular}
}

\caption{Comparison of learning rates (LR) when fine-tuning Mistral-7B model across language pairs, showing zero-shot and one-shot performance.}
\label{tab:lr-mistral}
\end{table}

\begin{table}[ht]
\centering
\resizebox{\columnwidth}{!}{%

\begin{tabular}{@{}lllcccc@{}}
\toprule
\textbf{Lang} & \textbf{Domain} & \textbf{LR} & \textbf{BLEU ↑} & \textbf{ChrF++ ↑} & \textbf{TER ↓} & \textbf{COMET ↑} \\
\midrule
\multirow{7}{*}{\textbf{EN-FR}} & \multirow{7}{*}{medical} & 1e-3 & 26.70 & 51.82 & 74.46 & 27.97 \\
& & 1e-4 & 43.40 & 66.46 & 50.99 & 74.40 \\
& & 2e-4 & 41.53 & 65.25 & 53.01 & 71.01 \\
& & 4e-5 & 42.27 & 66.70 & 50.40 & 75.01 \\
& & 5e-5 & \textbf{44.02} & \textbf{67.08} & \textbf{50.24} & \textbf{75.34} \\
& & 6e-5 & 42.85 & 66.86 & 50.33 & 75.16 \\
& & 7e-5 & 43.76 & 66.77 & 50.69 & 74.40 \\
\midrule
\multirow{6}{*}{\textbf{EN-SW}} & \multirow{3}{*}{medical} & 1e-3 & 20.94 & 46.24 & 72.65 & 66.03 \\
& & 1e-4 & \textbf{36.32} & \textbf{61.57} & \textbf{49.38} & 83.83 \\
& & 5e-5 & 36.30 & 61.31 & 49.48 & \textbf{83.99} \\
\cmidrule(l){2-7}
& \multirow{3}{*}{generic} & 1e-3 & 21.88 & 45.49 & 78.23 & 67.72 \\
& & 1e-4 & 38.08 & 59.57 & 52.53 & 84.70 \\
& & 5e-5 & \textbf{38.90} & \textbf{60.14} & \textbf{51.98} & \textbf{85.08} \\
\bottomrule
\end{tabular}
}

\caption{Comparison of learning rates (LR) when fine-tuning NLLB-3.3 model for zero-shot (sentence-level) translation.}
\label{tab:lr-nllb}
\end{table}

\subsection{Dataset Packing}

This section examines the performances of the models with and without packing. Packing dataset, initially introduced by \citep{Raffel2020-T5},  helps increase the training efficiency by enabling fine-tuning on more sequences. In this method, multiple short examples are packed in the same input sequence. Packing tokens into the maximum sequence length helps improve training efficiency. While packing is primarily intended to improve the training throughput, our experiments demonstrate it also improves model performance. In our experiments, we use the HuggingFace's SFTTrainer, which supports example packing.

In all the experiments in the main paper, we use \textit{packing=True} for the \textit{SFTTrainer}. Hence, we set \textit{add\_eos\_token} to \textit{False} as the trainer's dataset packing is True. By default, \textit{SFTTrainer} adds \textit{eos\_token} if dataset \textit{packing=True}, so there is no need to add it manually. If packing is False, then \textit{add\_eos\_token} should be set to True. The evaluation results are presented in Table \ref{tab:packing-comparison}.

\begin{table}[ht]
\centering
\resizebox{\columnwidth}{!}{%

\begin{tabular}{@{}llllcccc@{}}
\toprule
\textbf{Lang} & \textbf{Domain} & \textbf{Context} & \textbf{Packing} & \textbf{BLEU ↑} & \textbf{ChrF++ ↑} & \textbf{TER ↓} & \textbf{COMET ↑} \\
\midrule

\multirow{4}{*}{\textbf{EN-FR}} & \multirow{4}{*}{medical} & \multirow{2}{*}{zero-shot} & False & 34.75 & 61.45 & 65.37 & 67.74 \\
& & & True & \textbf{39.33} & \textbf{64.57} & \textbf{51.18} & \textbf{71.89} \\
\cmidrule(l){3-8}

& & \multirow{2}{*}{one-shot} & False & 17.53 & 53.53 & 195.57 & 37.93 \\
& & & True & \textbf{45.31} & \textbf{68.57} & \textbf{45.81} & \textbf{74.04} \\
\midrule

\multirow{4}{*}{\textbf{EN-PT}} & \multirow{4}{*}{medical} & \multirow{2}{*}{zero-shot} & False & 39.39 & 64.32 & 47.30 & 79.15 \\
& & & True & \textbf{40.05} & \textbf{64.68} & \textbf{47.08} & \textbf{80.12} \\
\cmidrule(l){3-8}
& & \multirow{2}{*}{one-shot} & False & 39.70 & 64.49 & 47.26 & 79.63 \\
& & & True & \textbf{40.39} & \textbf{64.86} & \textbf{47.11} & \textbf{80.61} \\
\midrule

\multirow{8}{*}{\textbf{EN-SW}} & \multirow{4}{*}{medical} & \multirow{2}{*}{zero-shot} & False & \textbf{18.04} & \textbf{44.15} & \textbf{81.44} & \textbf{56.30} \\
& & & True & 13.38 & 38.06 & 96.06 & 46.49 \\
\cmidrule(l){3-8}
& & \multirow{2}{*}{one-shot} & False & 12.27 & \textbf{42.26} & 121.31 & \textbf{54.45} \\
& & & True & \textbf{15.67} & 40.67 & \textbf{84.47} & 49.57 \\
\cmidrule(l){2-8}
& \multirow{4}{*}{generic} & \multirow{2}{*}{zero-shot} & False & \textbf{24.25} & \textbf{46.07} & \textbf{75.28} & \textbf{60.51} \\
& & & True & 19.12 & 41.71 & 94.07 & 52.85 \\
\cmidrule(l){3-8}
& & \multirow{2}{*}{one-shot} & False & 10.30 & 39.53 & 172.85 & \textbf{54.83} \\
& & & True & \textbf{21.12} & \textbf{43.16} & \textbf{85.21} & 54.37 \\
\midrule
\multirow{8}{*}{\textbf{SW-EN}} & \multirow{4}{*}{medical} & \multirow{2}{*}{zero-shot} & False & 30.12 & \textbf{52.96} & 63.28 & 77.12 \\
& & & True & \textbf{31.81} & 52.87 & \textbf{58.53} & \textbf{77.38} \\
\cmidrule(l){3-8}
& & \multirow{2}{*}{one-shot} & False & 9.81 & 41.21 & 252.60 & 67.72 \\
& & & True & \textbf{32.25} & \textbf{53.50} & \textbf{57.52} & \textbf{78.46} \\
\cmidrule(l){2-8}
& \multirow{4}{*}{generic} & \multirow{2}{*}{zero-shot} & False & 34.42 & \textbf{53.49} & 60.02 & \textbf{76.38} \\
& & & True & \textbf{34.49} & 52.93 & \textbf{58.24} & 75.87 \\
\cmidrule(l){3-8}
& & \multirow{2}{*}{one-shot} & False & 6.15 & 31.94 & 459.91 & 60.20 \\
& & & True & \textbf{34.97} & \textbf{53.28} & \textbf{57.21} & \textbf{76.06} \\
\bottomrule
\end{tabular}
}
\caption{Comparison of packing settings when fine-tuning Mistral.}
\label{tab:packing-comparison}
\end{table}

\section{Appendix: Inference}
\label{sec:appendix-inference}

\subsection{CTranslate2 vs vLLM}

CTranslate2 and vLLM are both efficient inference engines for LLMs. While they have an overlapping list of supported models, some models are not supported by both of them. For example, NLLB-200 models are supported by CTranslate2, but not supported by vLLM. At the time of writing this paper, Mixtral 8x7B is not supported by CTranslate2. For the sake of consistency of our results, all the models reported in the main paper use vLLM, except NLLB-200 3.3. Nevertheless, in this section, we show a brief comparison between CTranslate2 and vLLM evaluation results. The evaluation results are generally comparable between the two inference engines when evaluating Mistral and NLLB.

\begin{table}[ht]
\centering
\resizebox{\columnwidth}{!}{

\begin{tabular}{@{}llllcccc@{}}
\toprule
\textbf{Lang} & \textbf{Domain} & \textbf{Context} & \textbf{Engine} & \textbf{BLEU ↑} & \textbf{ChrF++ ↑} & \textbf{TER ↓} & \textbf{COMET ↑} \\
\midrule
\multirow{4}{*}{\textbf{EN-FR}} & \multirow{4}{*}{medical} & \multirow{2}{*}{zero-shot} & CT2 & 30.54 & 54.76 & 62.27 & \textbf{48.89} \\
& & & vLLM & \textbf{30.61} & \textbf{54.79} & \textbf{62.14} & 48.56 \\
\cmidrule(l){3-8}
& & \multirow{2}{*}{one-shot} & CT2 & \textbf{46.18} & \textbf{66.18} & \textbf{47.59} & \textbf{69.25} \\
& & & vLLM & \textbf{46.18} & 66.12 & 48.02 & 69.13 \\
\midrule
\multirow{4}{*}{\textbf{EN-PT}} & \multirow{4}{*}{medical} & \multirow{2}{*}{zero-shot} & CT2 & 33.53 & \textbf{60.09} & 55.39 & 63.22 \\
& & & vLLM & \textbf{33.60} & 60.08 & \textbf{55.23} & \textbf{63.52} \\
\cmidrule(l){3-8}
& & \multirow{2}{*}{one-shot} & CT2 & 37.63 & \textbf{62.92} & \textbf{49.26} & 77.99 \\
& & & vLLM & \textbf{37.64} & \textbf{62.92} & 49.35 & \textbf{78.05} \\
\midrule
\multirow{8}{*}{\textbf{EN-SW}} & \multirow{4}{*}{medical} & \multirow{2}{*}{zero-shot} & CT2 & \textbf{0.67} & 10.88 & 312.19 & \textbf{25.89} \\
& & & vLLM & 0.64 & \textbf{11.02} & \textbf{309.37} & 25.79 \\
\cmidrule(l){3-8}
& & \multirow{2}{*}{one-shot} & CT2 & 1.09 & 13.85 & 326.60 & 22.35 \\
& & & vLLM & \textbf{1.20} & \textbf{14.67} & \textbf{320.65} & \textbf{22.72} \\
\cmidrule(l){2-8}
& \multirow{4}{*}{generic} & \multirow{2}{*}{zero-shot} & CT2 & 1.72 & 12.74 & \textbf{289.67} & 29.64 \\
& & & vLLM & \textbf{1.82} & \textbf{13.15} & \textbf{289.67} & \textbf{30.15} \\
\cmidrule(l){3-8}
& & \multirow{2}{*}{one-shot} & CT2 & 2.49 & 15.96 & \textbf{320.44} & 27.19 \\
& & & vLLM & \textbf{2.70} & \textbf{17.02} & \textbf{320.44} & \textbf{27.88} \\
\midrule
\multirow{8}{*}{\textbf{SW-EN}} & \multirow{4}{*}{medical} & \multirow{2}{*}{zero-shot} & CT2 & 11.84 & \textbf{33.31} & 98.46 & \textbf{60.11} \\
& & & vLLM & \textbf{11.94} & 32.67 & \textbf{97.07} & 59.12 \\
\cmidrule(l){3-8}
& & \multirow{2}{*}{one-shot} & CT2 & \textbf{16.86} & \textbf{37.97} & \textbf{82.50} & \textbf{66.41} \\
& & & vLLM & 16.74 & 36.83 & 85.52 & 65.26 \\
\cmidrule(l){2-8}
& \multirow{4}{*}{generic} & \multirow{2}{*}{zero-shot} & CT2 & 16.77 & \textbf{36.99} & 95.29 & \textbf{60.03} \\
& & & vLLM & \textbf{17.04} & 36.69 & \textbf{93.06} & 59.83 \\
\cmidrule(l){3-8}
& & \multirow{2}{*}{one-shot} & CT2 & \textbf{21.83} & \textbf{40.62} & \textbf{79.75} & \textbf{63.53} \\
& & & vLLM & 21.44 & 40.16 & 80.47 & 63.27 \\
\bottomrule
\end{tabular}
}
\caption{Evaluation of Mistral 7B performance (baseline model without quantization) using CTranslate2 (CT2) and vLLM inference engines, showing comparable results.}
\label{tab:inference-comparison}
\end{table}

\begin{table*}[ht]
\centering
\resizebox{\textwidth}{!}{
\begin{tabular}{@{}llllcccclccc@{}}
\toprule
 &
   &
   &
   &
  \multicolumn{4}{c}{\textbf{NLLB-200 3.3B}} &
  \multicolumn{4}{c}{\textbf{Mistral 7B}} \\ 
  \cmidrule(lr){5-8}\cmidrule(lr){9-12}
  \addlinespace[3pt]
\textbf{Lang} &
  \textbf{Domain} &
  \textbf{Context} &
  \textbf{Quant.} &
  \textbf{BLEU ↑} &
  \textbf{ChrF++ ↑} &
  \textbf{TER ↓} &
  \textbf{COMET ↑} &
  \textbf{BLEU ↑} &
  \textbf{ChrF++ ↑} &
  \textbf{TER ↓} &
  \textbf{COMET ↑} \\ \midrule
\multirow{6}{*}{\textbf{EN-FR}} &
  \multirow{6}{*}{medical} &
  \multirow{3}{*}{zero-shot} &
  None &
  \textbf{41.37} &
  \textbf{64.10} &
  \textbf{52.11} &
  \textbf{59.32} &
  \textbf{30.54} &
  \textbf{54.76} &
  \textbf{62.27} &
  \textbf{48.89} \\
 &
   &
   &
  Float16 &
  41.30 &
  64.08 &
  52.15 &
  59.22 &
  30.47 &
  54.75 &
  62.28 &
  48.51 \\
 &
   &
   &
  Int8 &
  41.33 &
  64.07 &
  52.17 &
  59.09 &
  30.03 &
  54.18 &
  63.85 &
  47.00 \\ \cmidrule(l){3-12} 
 &
   &
  \multirow{3}{*}{one-shot} &
  None &
  \textbf{43.04} &
  \textbf{65.55} &
  54.49 &
  \textbf{55.84} &
  \textbf{46.18} &
  \textbf{66.18} &
  \textbf{47.59} &
  69.25 \\
 &
   &
   &
  Float16 &
  43.00 &
  65.54 &
  54.62 &
  55.83 &
  46.15 &
  66.16 &
  47.90 &
  \textbf{69.27} \\
 &
   &
   &
  Int8 &
  43.03 &
  65.31 &
  \textbf{54.41} &
  55.44 &
  44.71 &
  65.84 &
  48.33 &
  69.13 \\ \midrule
\multirow{6}{*}{\textbf{EN-PT}} &
  \multirow{6}{*}{medical} &
  \multirow{3}{*}{zero-shot} &
  None &
  \textbf{39.32} &
  \textbf{64.98} &
  \textbf{48.19} &
  81.77 &
  \textbf{33.53} &
  \textbf{60.09} &
  \textbf{55.39} &
  63.22 \\
 &
   &
   &
  Float16 &
  39.30 &
  64.97 &
  48.21 &
  81.81 &
  33.52 &
  60.06 &
  55.50 &
  \textbf{63.23} \\
 &
   &
   &
  Int8 &
  39.13 &
  64.89 &
  48.35 &
  \textbf{81.87} &
  32.62 &
  59.78 &
  56.59 &
  59.77 \\ \cmidrule(l){3-12} 
 &
   &
  \multirow{3}{*}{one-shot} &
  None &
  \textbf{39.79} &
  \textbf{64.36} &
  \textbf{48.38} &
  79.04 &
  \textbf{37.63} &
  \textbf{62.92} &
  \textbf{49.26} &
  77.99 \\
 &
   &
   &
  Float16 &
  \textbf{39.79} &
  \textbf{64.36} &
  \textbf{48.38} &
  \textbf{79.05} &
  37.63 &
  62.91 &
  49.34 &
  77.98 \\
 &
   &
   &
  Int8 &
  39.74 &
  64.33 &
  48.42 &
  78.95 &
  37.49 &
  62.82 &
  49.33 &
  \textbf{78.25} \\ \midrule
\multirow{12}{*}{\textbf{EN-SW}} &
  \multirow{6}{*}{medical} &
  \multirow{3}{*}{zero-shot} &
  None &
  32.61 &
  58.24 &
  53.29 &
  82.85 &
  \textbf{0.67} &
  \textbf{10.88} &
  312.19 &
  25.89 \\
 &
   &
   &
  Float16 &
  32.87 &
  \textbf{58.41} &
  53.08 &
  82.86 &
  0.65 &
  10.94 &
  315.29 &
  \textbf{26.10} \\
 &
   &
   &
  Int8 &
  \textbf{32.88} &
  58.36 &
  \textbf{53.08} &
  \textbf{83.03} &
  0.49 &
  10.22 &
  \textbf{290.16} &
  25.34 \\ \cmidrule(l){3-12} 
 &
   &
  \multirow{3}{*}{one-shot} &
  None &
  34.06 &
  59.12 &
  51.57 &
  82.37 &
  \textbf{1.09} &
  \textbf{13.85} &
  326.60 &
  \textbf{22.35} \\
 &
   &
   &
  Float16 &
  \textbf{34.07} &
  \textbf{59.23} &
  51.45 &
  82.38 &
  1.08 &
  14.11 &
  334.16 &
  22.30 \\
 &
   &
   &
  Int8 &
  33.97 &
  59.15 &
  \textbf{51.41} &
  \textbf{82.42} &
  0.97 &
  13.18 &
  \textbf{283.04} &
  21.87 \\ \cmidrule(l){2-12} 
 &
  \multirow{6}{*}{generic} &
  \multirow{3}{*}{zero-shot} &
  None &
  \textbf{35.57} &
  57.55 &
  \textbf{54.91} &
  83.90 &
  \textbf{1.72} &
  12.74 &
  303.84 &
  29.64 \\
 &
   &
   &
  Float16 &
  35.54 &
  \textbf{57.56} &
  54.93 &
  \textbf{83.92} &
  \textbf{1.72} &
  \textbf{12.83} &
  309.21 &
  \textbf{29.68} \\
 &
   &
   &
  Int8 &
  35.48 &
  57.46 &
  55.00 &
  83.85 &
  1.30 &
  11.50 &
  \textbf{278.97} &
  28.16 \\ \cmidrule(l){3-12} 
 &
   &
  \multirow{3}{*}{one-shot} &
  None &
  33.79 &
  55.93 &
  56.34 &
  83.23 &
  2.49 &
  15.96 &
  338.57 &
  27.19 \\
 &
   &
   &
  Float16 &
  \textbf{33.86} &
  \textbf{55.96} &
  56.32 &
  83.24 &
  \textbf{2.50} &
  \textbf{16.23} &
  344.52 &
  \textbf{27.35} \\
 &
   &
   &
  Int8 &
  33.73 &
  55.90 &
  \textbf{56.22} &
  \textbf{83.27} &
  1.99 &
  14.45 &
  \textbf{294.13} &
  25.65 \\ \midrule
\multirow{12}{*}{\textbf{SW-EN}} &
  \multirow{6}{*}{medical} &
  \multirow{3}{*}{zero-shot} &
  None &
  \textbf{44.58} &
  \textbf{64.73} &
  \textbf{42.71} &
  84.11 &
  \textbf{12.25} &
  \textbf{33.46} &
  \textbf{93.90} &
  60.06 \\
 &
   &
   &
  Float16 &
  44.48 &
  64.69 &
  \textbf{42.71} &
  \textbf{84.17} &
  12.19 &
  33.37 &
  95.38 &
  \textbf{60.18} \\
 &
   &
   &
  Int8 &
  44.28 &
  64.60 &
  43.03 &
  84.11 &
  11.15 &
  32.79 &
  102.77 &
  59.75 \\ \cmidrule(l){3-12} 
 &
   &
  \multirow{3}{*}{one-shot} &
  None &
  \textbf{43.66} &
  \textbf{64.03} &
  \textbf{42.85} &
  \textbf{83.92} &
  16.35 &
  37.66 &
  85.37 &
  66.24 \\
 &
   &
   &
  Float16 &
  \textbf{43.66} &
  64.01 &
  42.86 &
  83.89 &
  16.46 &
  37.67 &
  84.67 &
  \textbf{66.29} \\
 &
   &
   &
  Int8 &
  43.35 &
  63.72 &
  43.23 &
  83.78 &
  \textbf{16.99} &
  \textbf{37.72} &
  \textbf{82.32} &
  66.10 \\ \cmidrule(l){2-12} 
 &
  \multirow{6}{*}{generic} &
  \multirow{3}{*}{zero-shot} &
  None &
  42.87 &
  \textbf{61.59} &
  48.69 &
  82.00 &
  16.78 &
  \textbf{36.97} &
  95.5 &
  59.99 \\
 &
   &
   &
  Float16 &
  \textbf{42.89} &
  61.56 &
  48.63 &
  \textbf{82.01} &
  16.76 &
  36.92 &
  95.21 &
  \textbf{60.06} \\
 &
   &
   &
  Int8 &
  42.83 &
  61.53 &
  \textbf{48.59} &
  82.00 &
  \textbf{17.25} &
  36.82 &
  \textbf{92.17} &
  59.62 \\ \cmidrule(l){3-12} 
 &
   &
  \multirow{3}{*}{one-shot} &
  None &
  41.38 &
  59.99 &
  50.85 &
  80.92 &
  21.82 &
  40.63 &
  79.79 &
  \textbf{63.55} \\
 &
   &
   &
  Float16 &
  \textbf{41.40} &
  \textbf{60.01} &
  50.89 &
  80.95 &
  \textbf{22.03} &
  \textbf{40.75} &
  \textbf{79.05} &
  63.54 \\
 &
   &
   &
  Int8 &
  41.35 &
  \textbf{60.01} &
  \textbf{50.72} &
  \textbf{80.97} &
  21.50 &
  40.17 &
  80.12 &
  63.12 \\ \bottomrule
\end{tabular}
}
\caption{Evaluation of the translation performance of \textbf{NLLB-200 3.3B} and \textbf{Mistral 7B}, without quantization and with lower-precision quantization formats (float16 and int8) using CTranslate2. The results indicate similar performance across all formats.}
\label{tab:quantization}
\end{table*}

\subsection{Quantization}
\label{sec:quantization}

We compared quantization formats when working with NLLB-200 CTranslate2, i.e. int8, float16, no quantization (cf.~Table \ref{tab:quantization}). Models with int8 quantization have the smallest size, and they excel on CPU setups, while models with float16 quantization are smaller than non-quantized models, but larger than int8 models. Float16 models are faster on GPU setups. When it comes to translation performance, the differences between the three setups are marginal. All NLLB-200 3.3B results in the main paper are for non-quantized models.

Moreover, we compared generation with Llama-3 70B with and without Activation-aware Weight Quantization (AWQ) using vLLM. As Table \ref{tab:quantization-vllm} shows, results without quantization are slightly better. All the results in the main paper for LLMs are for non-quantized models, except Llama-3.1 405B and DeepSeek V3 685B, due to computing constrains. Hence, we tested Llama-3 70B with and without quantization and reported the results in this section to assess how AWQ affects the quality.

\begin{table}[!ht]
\centering
\resizebox{\columnwidth}{!}{
\begin{tabular}{@{}lllcccc@{}}
\toprule
\textbf{Lang} & \textbf{Context} & \textbf{Quant.} & \textbf{BLEU ↑} & \textbf{ChrF++ ↑} & \textbf{TER ↓} & \textbf{COMET ↑} \\
\midrule

\multirow{4}{*}{\textbf{EN-FR}} & \multirow{2}{*}{zero-shot} 
& None & \textbf{39.74} & \textbf{62.51} & \textbf{53.36} & \textbf{66.75} \\
& & AWQ & 39.02 & 61.97 & 54.69 & 64.97 \\
\cmidrule(l){2-7}
& \multirow{2}{*}{one-shot} 
& None & \textbf{51.93} & \textbf{70.57} & 43.03 & \textbf{76.76} \\
& & AWQ & 51.83 & 70.43 & \textbf{43.00} & 76.19 \\
\midrule

\multirow{4}{*}{\textbf{EN-PT}} & \multirow{2}{*}{zero-shot} 
& None & \textbf{43.86} & \textbf{67.35} & \textbf{43.95} & \textbf{84.05} \\
& & AWQ & 43.12 & 66.59 & 45.58 & 82.12 \\
\cmidrule(l){2-7}
& \multirow{2}{*}{one-shot} 
& None & \textbf{45.95} & \textbf{68.82} & \textbf{41.68} & \textbf{86.58} \\
& & AWQ & 45.59 & 68.57 & 42.12 & 86.10 \\

\bottomrule
\end{tabular}
}
\caption{Comparison of translation performance of \textbf{Llama-3 70B}, with and without AWQ quantization, using vLLM.}
\label{tab:quantization-vllm}
\end{table}

\section{Appendix: Other Models}
\label{sec:appendix-models}

\subsection{Llama 3.3 vs. Llama 3}

We evaluated the \textit{Llama-3.3 70B} instruction-tuned model, as it was promoted to be more efficient with “enhanced” quality (cf.~the official model card). However, our evaluation does not show any considerable improvement compared to Llama-3 70B (cf.~Table \ref{tab:llama3.3}), probably due to the different types of tasks they target.

\begin{table}[ht]
\centering
\resizebox{\columnwidth}{!}{

\begin{tabular}{@{}lllcccc@{}}
\toprule
\textbf{Lang} & \textbf{Model} & \textbf{Context} & \textbf{BLEU ↑} & \textbf{ChrF++ ↑} & \textbf{TER ↓} & \textbf{COMET ↑} \\
\midrule

\multirow{4}{*}{\textbf{EN-FR}} & \multirow{2}{*}{Llama-3 70B} 
& zero-shot & 39.74 & 62.51 & 53.36 & 66.75 \\
& & one-shot & \textbf{51.93} & \textbf{70.57} & \textbf{43.03} & 76.76 \\
\cmidrule(l){2-7}
& \multirow{2}{*}{Llama-3.3 70B} 
& zero-shot & 39.73 & 62.99 & 55.22 & 69.16 \\
& & one-shot & 48.72 & 69.53 & 48.17 & \textbf{76.98} \\
\midrule

\multirow{4}{*}{\textbf{EN-PT}} & \multirow{2}{*}{Llama-3 70B} 
& zero-shot & 43.86 & 67.35 & 43.95 & 84.05 \\
& & one-shot & \textbf{45.95} & \textbf{68.82} & \textbf{41.68} & \textbf{86.58} \\
\cmidrule(l){2-7}
& \multirow{2}{*}{Llama-3.3 70B} 
& zero-shot & 38.24 & 64.56 & 54.99 & 78.2 \\
& & one-shot & 45.25 & 68.42 & 42.6 & 86.28 \\

\bottomrule
\end{tabular}
}
\caption{Comparison of translation performance of Llama-3 70B and Llama-3.3~70B instruct. In general, Llama-3 70B shows better performance.}
\label{tab:llama3.3}
\end{table}

\end{document}